\documentclass{article}

\usepackage{arxiv}
\pdfoutput=1 
\usepackage[utf8]{inputenc} 
\usepackage[T1]{fontenc}    
\usepackage{hyperref}       
\usepackage{url}            
\usepackage{booktabs}       
\usepackage{amsfonts}       
\usepackage{nicefrac}       
\usepackage{microtype}      
\usepackage{lipsum}

\title{Selective Network Discovery via Deep Reinforcement Learning on Embedded Spaces}

\author{
 Peter Morales \thanks{corresponding author} \\
 MIT Lincoln Laboratory\\
 Lexington, MA\\
 \texttt{petermor@mit.edu} \\
 \And
 Rajmonda Sulo Caceres \\
 MIT Lincoln Laboratory\\
 Lexington, MA\\
 \texttt{rajmonda.caceres@ll.mit.edu} \\
 \And
 Tina Eliassi-Rad \\
 Northeastern University\\
 Boston, MA\\
 \texttt{t.eliassirad@northeastern.edu} \\
}
\usepackage{graphicx}
\usepackage{epstopdf}
\usepackage[caption = false]{subfig}
\usepackage{url}

\usepackage{svg}
\usepackage{booktabs}
\usepackage{makecell}
\usepackage{multicol}

\usepackage{commath}
\usepackage{amsmath}
\usepackage{fancyhdr,amssymb}
\usepackage[ruled,vlined,linesnumbered]{algorithm2e}
\include{pythonlisting}

\begin{document}
\newcommand{\Gt}{\ensuremath{G_\text{true}}}
\newcommand{\Ct}{\ensuremath{C_\text{true}}}
\newcommand{\Nt}{\ensuremath{\mathcal{V}_\text{true}}}
\newcommand{\Et}{\ensuremath{\mathcal{E}_\text{true}}}
\newcommand{\Comt}{\ensuremath{\mathcal{C}_\text{true}}}

\newcommand{\Gs}{\ensuremath{G_\text{samp}}}
\newcommand{\Ns}{\ensuremath{\mathcal{V}_\text{samp}}}
\newcommand{\Es}{\ensuremath{\mathcal{E}_\text{samp}}}
\newcommand{\Coms}{\ensuremath{\mathcal{C}_\text{samp}}}

\newcommand{\target}{\ensuremath{\text{target}}}

\maketitle

\begin{abstract}
    
Complex networks are often either too large for full exploration, partially accessible, or partially observed. Downstream learning tasks on these incomplete networks can produce low quality results. In addition, reducing the incompleteness of the network can be costly and nontrivial. As a result, network discovery algorithms optimized for specific downstream learning tasks given resource collection constraints are of great interest. In this paper, we formulate the task-specific network discovery problem in an incomplete network setting as a sequential decision making problem. Our downstream task is selective harvesting,  the optimal collection of vertices with a particular attribute. We propose a framework, called Network Actor Critic (NAC), which learns a policy and notion of future reward in an offline setting via a deep reinforcement learning algorithm. The NAC paradigm utilizes a task-specific network embedding to reduce the state space complexity. A detailed comparative analysis of popular network embeddings is presented with respect to their role in supporting offline planning. Furthermore, a quantitative study is presented on several synthetic and real benchmarks using NAC and several baselines. We show that offline models of reward and network discovery policies lead to significantly improved performance when compared to competitive online discovery algorithms. Finally, we outline learning regimes where planning is critical in addressing sparse and changing reward signals.        

\end{abstract}

\section{Introduction}
Complex networks are critical to many applications such as those in the social, cyber, and bio domains. We commonly have access to partially observed data. The challenge is to discover enough of the complex network so that we can perform a learning task well. The network discovery step is especially critical in the case when the learning task has the characteristics of the ``needle in a haystack" problem. If the discovery process is not carefully tuned, the noise introduced, almost always, overwhelms the signal. This presents an optimization problem: how should we grow the incomplete network to achieve a learning objective on the network, while at the same time minimizing the cost of observing new data? 

In this work we view the network discovery problem from a decision theoretic lens, where notions of utility and resource cost are naturally defined and jointly leveraged in a sequential, closed-loop manner.  
In particular, we will leverage Reinforcement Learning (RL) and its mathematical formalism, Markov Decision Processes (MDP), a general decision theoretic model that allows us to treat network discovery as an interactive, sequential learning and planning problem. 
MDP approaches have been successfully used in many other application settings~\cite{Mnih2015,Heess2017,Silver2017}. However, the use of decision theoretic approaches in the context of discovery of complex networks is novel and presents very interesting research opportunities. In particular, it requires learning effective models of reward that can capture properties of network structure at various topological scales and learning contexts. The network science community has defined many such topological and task quality metrics, but, to-date, they have not been leveraged in the context of guiding the process of discovery of a partially observed, incomplete network. We consider the task of selective harvesting on graphs~\cite{Murai2017}, where the learning objective is to maximize the collection of nodes of a particular type, under budget constraints. We make the following contributions:
\renewcommand{\labelitemi}{$\bullet$}
\begin{itemize}
   \item We introduce a deep RL framework for task-driven discovery of incomplete networks. This formulation allows us to learn offline-trained models of environment dynamics and reward.
   \item We show that, for a variety of complex learning scenarios, the added feature of learning from closely related scenarios leads to substantial performance improvements relative to existing online discovery methods.
   \item We show that network embedding can play an important role in the convergence properties of the RL algorithm. It does so by imposing structure on the network state space and prioritizing navigation over this space.
   \item Among a class of embedding algorithms, we identify personalized Pagerank (PPR) as a suitable network embedding algorithm for the selective harvesting task. Our combined approach of PPR embedding and offline planning achieves substantial reductions in training and convergence time.
   \item Leveraging several evaluation metrics, we delineate learning regimes where embedding alone stops being effective and planning is required.
   \item Our approach is able to generalize well to unseen real network topologies and new downstream tasks.
\end{itemize}

\section{Related Work}
Our learning task falls under the category of finding the largest number of a particular type of node under budget constraints. The node type can be specified by the node attributes (for example, {\it follower} nodes on a twitter network), or they can be determined by node's participation on a particular  class of behavior (for example, membership to anomalous activity). Unlike the problem setting in~\cite{Wang2013}, we do not assume access to the full topology of the network and therefore have to perform the learning task with partial information.

Discovering incomplete networks with limited resources has received a lot of attention in recent literature. The primary learning objective in these works is to increase the visibility of the network topology by either increasing the number of undiscovered nodes~\cite{Larock2018,Soundarajan2015,Soundarajan2016},  or by increasing network coverage~\cite{Avrachenkov2014}. Our problem setting is the most similar to selective harvesting~\cite{Murai2017}. Our approach differs from~\cite{Murai2017} by leveraging the Reinforcement Learning paradigm to estimate offline models of network discovery strategies (policy) and node utility (reward) that are state-aware. More specifically, our approach explicitly connects the utility of a discovery choice to the network state when that choice was made. 

Reinforcement learning for tasks on complex networks is a relatively new perspective. Work in~\cite{Ho2015,Goindani2019} leverages Reinforcement Learning to engineer diffusion processes in networks assumed to be fully observed, while authors in~\cite{Mofrad2019} focus on the problem of graph partitioning. You et al.~\cite{You2018} leverage Reinforcement Learning to generate novel molecular graphs with desired domain-specified properties. There are connections to our problem setting. The graph generation is approached in a similar fashion to the network discovery problem, by iteratively expanding a seed graph via defined actions. There are, however, some important differences with our work. Since the application in~\cite{You2018} is molecular design, the size of the graphs they consider is very small. Their definition of reward and environment dynamics is tailored to the biochemical domain. Our approach is more general and can support discovery of different types of networks and different network sizes. Our notion of reward is also more general in that we do not utilize domain-specific properties to guide the learning process. 

De Cao and Kipf~\cite{DeCao2018} similarly to~\cite{You2018} focus on small molecular graph generation, and furthermore, they do not consider the generation process as a sequence of actions. 
Finally,~\cite{Dai2017,Mittal2019} leverage deep Reinforcement Learning techniques to learn a class of graph greedy optimization heuristics on fully observed networks.

\section{Problem Definition}
We start with the assumption that a network contains a target subnetwork representing a set of relevant vertices. The objective is to strategically explore and expand the network so that we optimize discovery of these relevant vertices. The decision making agent is initially given partial information about the network $G_0=(N_0,E_0)$. A subset of those vertices have their relevance status $C_0$ revealed as well, with $0$ representing non-target vertices and $1$ representing target vertices. We assume our exploration starts from a seed vertex belonging to the partial target subnetwork. At each step, the agent can choose from a set of vertices that are observed, but whose label is unknown. We refer to this set of vertices as the boundary set $\mathcal{B}$. After selecting a vertex, the agent can gain knowledge of the vertex label, as well as the identity of all its neighbors. An immediate reward is given if the selected vertex belongs to the target subnetwork. 

This problem may be stated as a Markov Decision Process (MDP). An MDP is defined by the tuple $\langle \mathcal{S}, \mathcal{A}, T, R, \gamma \rangle$:
\begin{figure}[!ht]
\begin{center}
\includegraphics[width=0.4\textwidth]{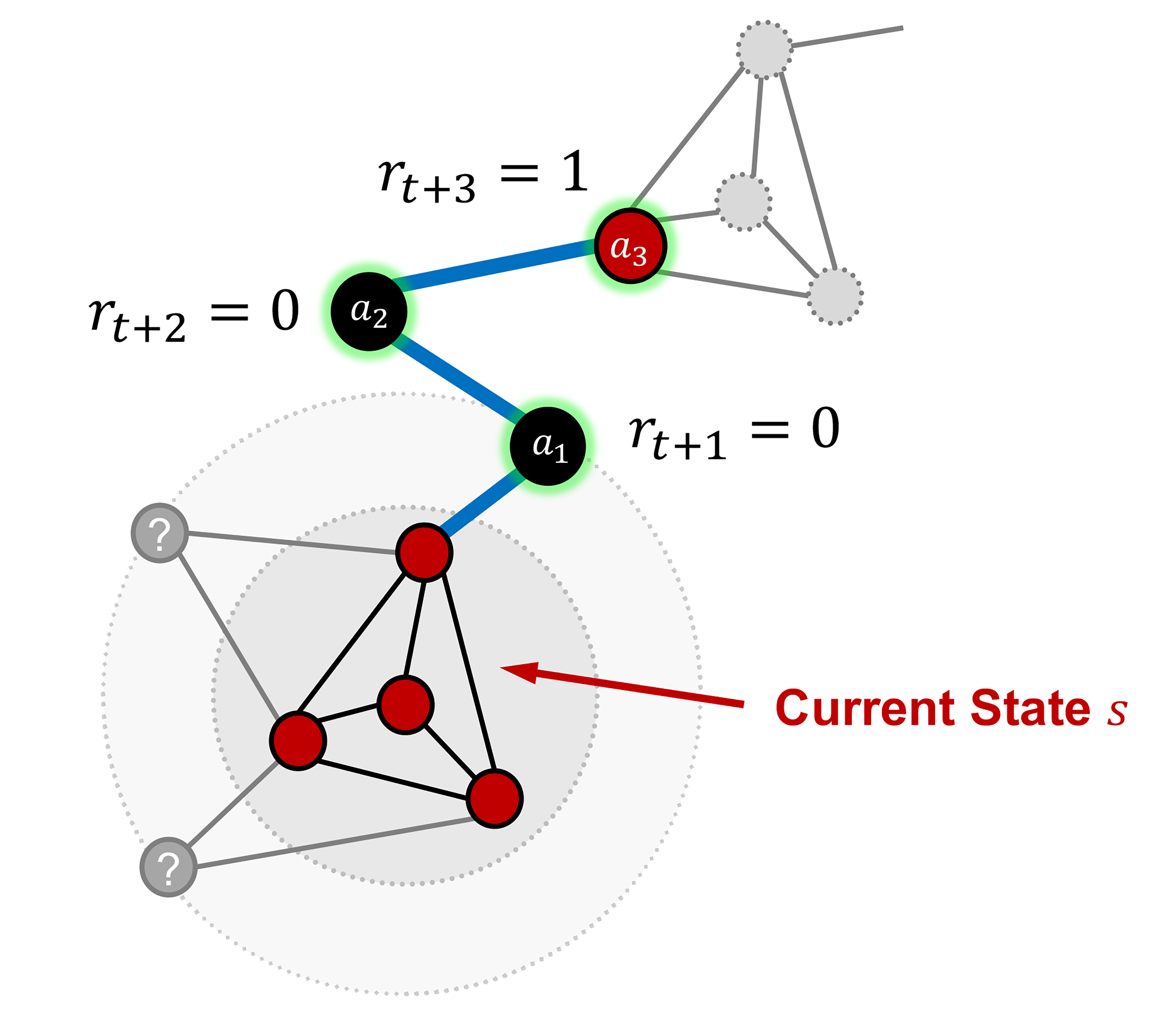}
\caption{Illustration of estimation of cumulative reward of state $s$ over a trajectory of length $h=3$, and discount factor $\gamma=0.5$; red nodes represent the node type we would like to discover:  $Q(s,a_1)= 1*0+1/2*0+1/4*1=1/4$.}
\label{fig:r_est}
\end{center}
\end{figure} 

\begin{itemize}
    \item The \textbf{state space}, $\mathcal{S}=\{s_t\}$, is the set of intermediate discovered networks. 
    \item The \textbf{action space}, $\mathcal{A}=\{A_t\}$, where $A_t=\{a \}$ is the set of boundary vertices at step $t$.
    \item The \textbf{transition model}, $T(s,a,s')=P(s'|s,a)$ encodes how the network state changes by specifying the probability of state $s$ transitioning to $s'$ given action $a$. We do not model this transition function explicitly and take the model-free approach, where we iteratively define and approximate reward without having to directly specify the network state transition probabilities. We make this more precise in Section~\ref{sec:alg}.
    \item The local \textbf{reward function}, $R(s_t, a_t)$ returns the reward gained by executing action $a$ in state $s$ and is defined as: $   R(s_t, a_t) = 1$ if $C(a_t)=1$. The total cumulative, action-specific reward, also referenced as the action-value function $Q$, is defined as:
    \begin{equation}
    \label{val}
Q(s,a)=[\sum_{t=0}^h \gamma^t R_{t+1}|s,a]
    \end{equation}
 with $\gamma$ representing a discount factor that captures the utility of exploring future graph states. Figure~\ref{fig:r_est} gives a simple illustration of how this cumulative reward is computed over a network topology. In the next section, we describe in detail our deep reinforcement learning algorithm.
 \end{itemize}

\section{Network Actor Critic (NAC) Algorithm}
\label{sec:alg}
\begin{figure}[ht]
\begin{center}
\includegraphics[width=0.50\textwidth]{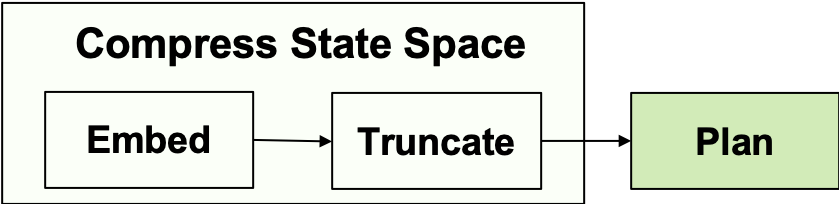}
\caption{Schematic approach of NAC algorithm. NAC uses a network embedding and truncation step to avoid an explosion in the state-action space as the network grows. The truncation block ensures a constant size input into the learned policy.}
\label{fig:nac-paradigm}
\end{center}
\end{figure}

Our network discovery algorithm has two main components as illustrated in Figure~\ref{fig:nac-paradigm}. The first component, ``Compress State Space", is concerned with effective ways of representing the large network state space so that policy learning can happen efficiently and optimally relative to our selective harvesting task.
The second component, ``Plan", utilizes the reinforcement learning framework and offline training to learn task-driven discovery strategies. We discuss both components in detail in the rest of this section.
\subsection{Compression of Network State Space}
\begin{figure}[ht]
\begin{center}
\includegraphics[width=0.99\textwidth]{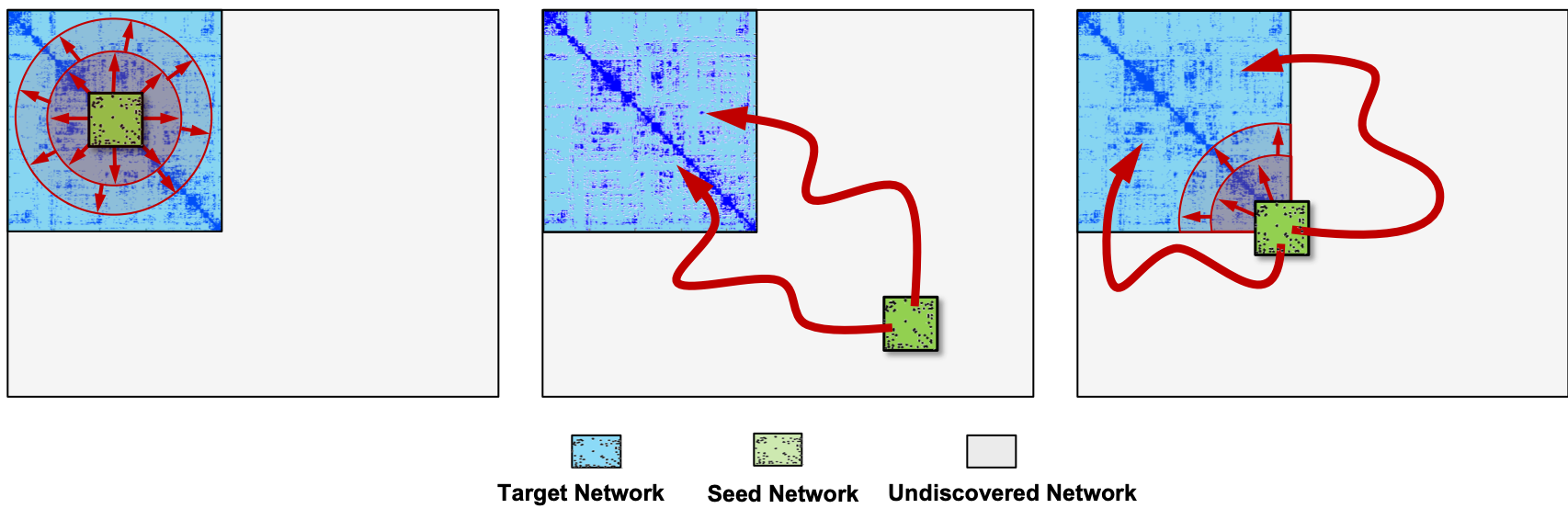}
\caption{Representative high level abstractions of network states for selective harvesting task.}
\label{fig:canonical-cases}
\end{center}
\end{figure}

Training an effective network discovery agent implies exploration over an extremely large network space. 
 However, not all the variation observed contributes to learning better discovery policies. In fact, for the task of selective harvesting, we can identify three representative, higher-level abstractions of the network states. As illustrated in 
Figure~\ref{fig:canonical-cases}, in the first canonical case, discovery starts within the region of interest and many of the relevant nodes we need to discover are nearby. In networks states that are similar to this canonical case, the optimal discovery agent would follow localized paths and primarily exploit rather then explore new regions. 
In the second canonical case, discovery starts outside the region of interest and the agent has to now explore longer, deeper paths in order to reach the target region. Finally, there is a hybrid canonical case, where discovery can start in the boundary of the target region and the agent has to more carefully decide when to exploit or explore. 


We consider various popular network embedding approaches including walk-based algorithms~\cite{Groover2016,Haveliwala2003,Murai2017} and matrix factorization algorithms~\cite{Torres2019,Pearson1901,Belkin2003} with the goal of collapsing network states into canonical representations. The embedding step learns a new similarity function between nodes in a network. Since our downstream task is selective harvesting from a seed node, we reorder the rows of the original adjacency matrix based on the new learned distance from the seed node, with closer nodes being ranked higher. The reordering step makes sure the discovery algorithm observes a prioritized set of boundary nodes. For additional efficiency gains, we truncate the reordered adjacency matrix and only retain the network defined by the top $k$ vertices. $k$ is a parameter we select and it defines the supporting network for computing potential discovery trajectories and long-term reward. 

In section~\ref{sec:embed}, we evaluate in detail the various embedding algorithms considered and identify the role they play in supporting the planning component of NAC.  Among the embedding algorithms we study, we identify personalized Pagerank (PPR)~\cite{Haveliwala2003} as performing the best in supporting policy learning for selective harvesting. 


\subsection{Offline Learning and Policy Optimization}
In our setting, learning of discovery strategies happens offline over a training set of possible discovery paths. We use simulated instances of both background networks and target subnetworks to generate paths or trajectories $\tau_h$ over the network state space. 

Each path $\tau_h$ represents an alternating sequence of discovered graph, action $\langle s_0,a_0,s_1,a_1,\ldots,a_h,s_h\rangle$, taken over $h$ steps. Since in this setting we have access to the ground truth vertex labels, we can map each discovery path to the corresponding cumulative reward value using equation (\ref{val}). An illustration is given in Figure~\ref{fig:r_est}.

Given the sampled trajectories, one of our learning objectives becomes to approximate the action-value function by minimizing the loss $L_{Q}(\phi)$,
\begin{equation}\label{loss}
L_{Q}(\phi)=||y_t - Q_\phi(x_t)||_2^2. 
\end{equation}
We formulate this objective by taking the input tuples of discovered graphs $s_t$, boundary nodes $a_t$  and corresponding cumulative reward values $Q_t$, such that $\langle x_t=(s_t,a_t),y_t=Q_t \rangle$. The approximated function $Q_\phi$ can then be utilized to estimate the policy function $\pi_\theta$, which defines the action probability distribution at each state. In particular, we estimate the advantage of choosing one node versus another at state $s_i$, 
\begin{equation}\label{adv}
\hat{A}_{t}=Q_\phi(s_t,a_t)-\sum_{a\in\mathcal{A}}Q_\phi(s_t,a).    
\end{equation}

This advantage is used to scale the policy gradient estimator, typically defined as, $\hat{g}_t=\hat{\mathbb{E}}_t\big[\hat{A}_t \nabla_\theta \text{log}\ \pi_\theta  \big].$ We utilize a proximal policy optimization (PPO) method \cite{Schulman2017} in order to compute this gradient. PPO methods are widely utilized for policy network optimization and have been demonstrated to achieve state of the art performance on graph tasks \cite{You2018}. The objective function utilized is defined in equation \ref{ppo},
\begin{equation}\label{ppo}
    L^{CLIP}(\theta)=\hat{\mathbb{E}}_t\big[\text{min}\big(\frac{\pi_\theta}{\pi_{\theta_{old}}}\hat{A}_t,\text{clip}(\frac{\pi_\theta}{\pi_{\theta_{old}}},1-\epsilon,1+\epsilon)\hat{A}_t\big) \big].
\end{equation}
Here, $\epsilon$ is used to bound the loss function and help with convergence. During offline training, we modify this objective to encourage exploration and reduce the number of required training epochs to converge to a solution. For equation \ref{objective}, $S$ denotes the entropy of policy $\pi_\theta$ in state $s_t$ and $c$ is used to balance exploitation vs exploration, 
\begin{equation}\label{objective}
L^{CLIP+S}=\hat{\mathbb{E}}_t\big[L^{clip}_t(\theta)+cS[\pi_\theta](s_t)\big].
\end{equation}
Both learning objectives (\ref{loss}) and (\ref{objective}) are jointly optimized via an actor critic training framework. This framework is detailed further below in the description of the \textit{Network Actor Critic} (NAC) algorithm. To help with training times, multiple instantiations of agents are run simultaneously. Collected $\{s_t,a_t,Q_t\}$ values are gathered from each agent and are stored in a buffer $\beta$ which is used to compute the losses for the value function and policy networks after a fixed time window of $T$ steps. 

\begin{algorithm}[ht]
\SetAlgoLined
 set hyper-parameters: exploration constant $c$, learning rate $\epsilon$, embedding function $e$, update window size $T$\;
 initialize: policy parameters $\theta$, value function parameters $\phi$, buffer $\beta$\;
 $\theta_{old}=\theta$\;
 \For{\text{t=1,2,...}}{
    $s \leftarrow e(G_{observed})$\;
    \For{\text{agent=1,2,...,N}}{
        $a\sim\pi_{\theta_{old}}$\;
        $r \leftarrow$ take action $a$ and save reward $r$\;
        $s' \leftarrow e(G_{observed})$\;
        $\beta \leftarrow$ save $(s_{t},a_{t},r_{t},s_{t+1})$ to buffer $\beta$\;
        $s \leftarrow s'$
    }
    \If{\text{t modulo T is 0}}{
    Compute batch update tuples $\{s_t,a_t,Q^\pi_t\}$ over horizon $H$ using  $\beta$\;
    Batch update $\phi$ via $\nabla_\phi L^{Q}(\phi)$ using eq. (\ref{loss})\;
    Compute $\hat{A}_t$ using eq. (\ref{adv})\;
    Batch update $\theta$ via $\nabla_\theta L^{CLIP+S_c}(\theta)$ using eq. (\ref{objective})\;
    }
 }
 \caption{Network Actor Critic (NAC)}
 \label{algorithm}
\end{algorithm}

\subsubsection{Training and Network Details}
The NAC algorithm is updated differently during offline training versus online evaluation. During offline training, the ADAM optimizer \cite{Kingma2014} is used to update network parameters $\mathbf{\theta}$ and $\mathbf{\phi}$ for the policy and value function networks. In offline training, eight agents simultaneously carry out the anomaly discovery task on a unique network realization generated using the random graphs outlined in Table \ref{tab:syndata}. During offline training, the hyper parameters used are: $T=32$, $H=4$, $c=0.2$, $\epsilon=0.1$, $\gamma=0.1$, and learning rate $\lambda=1e-4$. For online evaluation, we used a single agent and parameters $T=1$, $H=1$, $\gamma=1$, $\epsilon=0.2$, $c=0$, and $\lambda=1e-3$. The policy and value function networks are both comprised of 3 convolutional layers with 64 hidden channels and a final fully connected layer.

\section{NAC Performance Results}
We evaluate our algorithm against several learning scenarios for both synthetic and realistic datasets. Next we describe our datasets and baselines used for comparison.
\subsection{Datasets}
\paragraph{Synthetic Datasets:} We approach synthetic graph generation by individually modeling a background network (i.e., the network that does not contain any of the target nodes), and the foreground network (i.e., the network that only contains the target nodes and the interactions among them). We use two models to generate samples of background networks. \textit{Stochastic Block model} (SBM)~\cite{Holland1983} is a common generative graph model that allows us to model community structure as dense subgraphs sparsely connected with the rest of the network. \textit{Lancichinetti--Fortunato--Radicchi} (LFR) model~\cite{Lancichinetti2008} is another frequently used generative model that, in contrast to SBM, allows us to simulate network samples with skewed degree distributions and skewed community sizes, and therefore is able to capture more realistic and complex properties of real networks. Finally, we use the \textit{Erd\H{o}s-Renyi} (ER) model~\cite{Holland1983} to simulate the foreground network. ER is a simple generative model where vertices are connected with equal probability $p_{f}$ controlling the density of the foreground network. Parameter choices for all the models above are detailed in Table~\ref{tab:syndata}. 

In order to create a background plus foreground network sample, we select a subset of the nodes from the background network that will represent the identity of the target nodes. We then simulate an ER subnetwork on these nodes and replace their background induced subnetwork with the ER subnetwork. We reference this process in the rest of the paper as \textit{embedding} the foreground subnetwork.

\paragraph{Real Datasets:}
We analyzed two Facebook datasets~\cite{Roz2018} representing pages of different categories as nodes and mutual likes as edges. For both cases, we study the discovery of a target set of vertices, where we control how we generate and embed them in the background network. In particular, we embed a synthetic foreground subnetwork consisting of a denser (anomalous) ER graph with size $n_f=80$ and density $p_f=0.003$. 
We also consider the Livejournal dataset~\cite{Murai2017}. This dataset represents an online social network with users representing the nodes, and their self-declared friendships representing the edges. For each user, there is also information on the groups they have joined. Similarly to~\cite{Murai2017}, we use one of the listed groups as the target class. The Livejournal dataset represents a departure from the two Facebook datasets, both in terms of its much larger size, but also because the target class does not represent an anomaly. A few topological characteristics of the real networks described here, as well as details on their target class are listed in Table~\ref{tab:rdata}.

\subsection{Baselines}
We evaluate the NAC algorithm by comparing performance with two top performing online network discovery approaches. The \textit{Network Online Learning }(NOL)~\cite{Larock2018} algorithm learns an online regression function that maximizes discovery of previously unobserved nodes for
a given number of queries. We modify the objective of NOL to match our problem setting by requiring the discovery of previously unobserved nodes of a particular type. A second baseline we consider is the \textit{Directed Diversity Dynamic Thompson Sampling} ($D^3TS$)~\cite{Murai2017} approach. $D^3TS$  is  stochastic multi-armed bandit approach that leverages different node classifiers and Thompson sampling to diversify the selection of a boundary node. 
We also compare to a simple fixed node selection heuristic referenced in~\cite{Murai2017} called \textit{Maximum Observed Degree} (MOD). At every decision step, MOD selects the node with the highest number of observed neighbors that have the desired label. Finally, we compare to the heuristic that at each step selects the node with the highest PPR score to show the additional benefit provided from NAC.
\begin{table}[ht]\small
\begin{center}
\begin{tabular}{l c c}
\toprule
\textbf{Model} & \textbf{Type} & \textbf{Parameters} \\
\toprule
SBM & Background & $k=[1,10], p_i=[0.01,0.4],r=[0.005,0.25],i=1 \ldots k$ \\
\hline
LFR & Background & \makecell{$\tau_1=[3,2], \tau_2=(1,1.9], \mu=[0.1,0.4], \langle d \rangle=[32,256],$\\
$d_{\max}=[256,2048],\min_{c}=[256,1000], \max_{c}=[512,2000]$} \\
\hline
ER  & Foreground     & $n_f=\{30,40,80\},k_f=\{1,2,4\},p_f=[0.5,1]$\\
\hline
\end{tabular}
\end{center}
\caption{Detailed list of parameter values used for synthetic networks. Number of vertices is represented by $N=4000$. SBM parameters are: $k$ represents the number of communities, $p_{i}$ the within-community edge probability for community $i$, $r$ the across-community edge probability, such that $p_{i} > r$. LFR parameters are: $\tau_1,\tau_2$ skewness parameters for degree and cluster size distributions respectively, $\langle d \rangle$ represents the average network degree, $d_{\min},d_{\max}$ represent the min and max values of degree distribution, $\min_c$ and $\max_c$ represent the sizes of smallest and largest clusters, and finally $n_f,k_f,p_f$ represent the size of the foreground subnetwork, number of foreground subnetworks and its edge probability, respectively.}
\label{tab:syndata}

\begin{center}
\begin{tabular}{p{3cm} p{2cm} p{2cm} p{2cm} p{2cm}}
\toprule
 \textbf{Name} & \textbf{\# Nodes}  & \textbf{\# Edges} & \textbf{Target Type}  & \textbf{Target Size} \\
 \toprule
 Facebook Politician  &   5,908 & 41,729  &  Synthetic & 80 \\ 
 Facebook TV Shows  &   3,892  & 17,262 &  Synthetic & 80 \\
 Livejournal  &  $\approx$ 4,000k   & $\approx$ 35,000k & Real & $\approx$ 1,400
\end{tabular}
\end{center}
\caption{Characteristics of the real networks and corresponding target classes.}
\label{tab:rdata}
\end{table}

\begin{figure}[ht]
\begin{center}
\subfloat[Easier target detectability \label{fig:high-snr}]{
\includegraphics[width=0.36\linewidth]{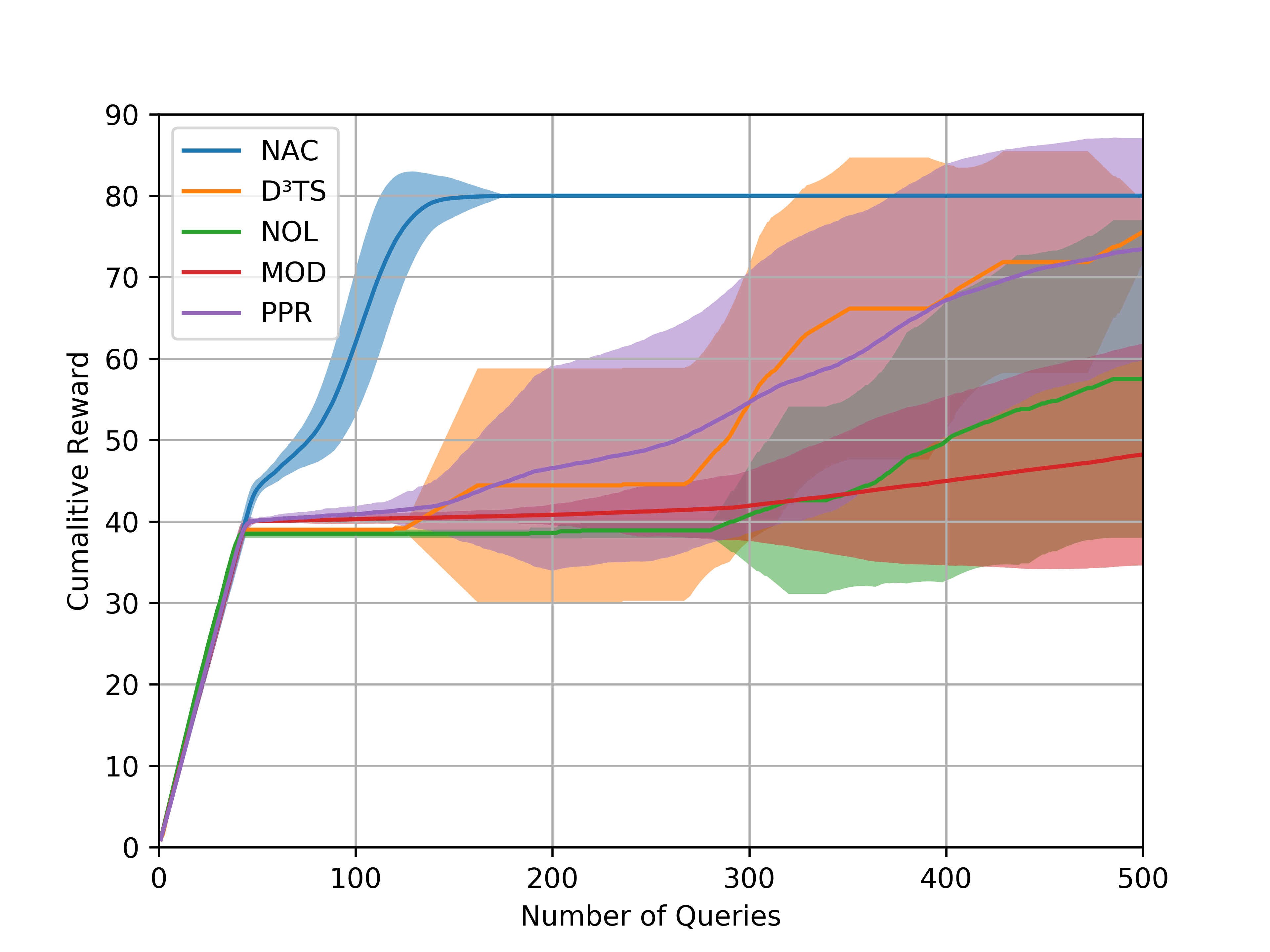}
}
\subfloat[Harder target detectability \label{fig:low-snr}]{
\includegraphics[width=0.36\linewidth]{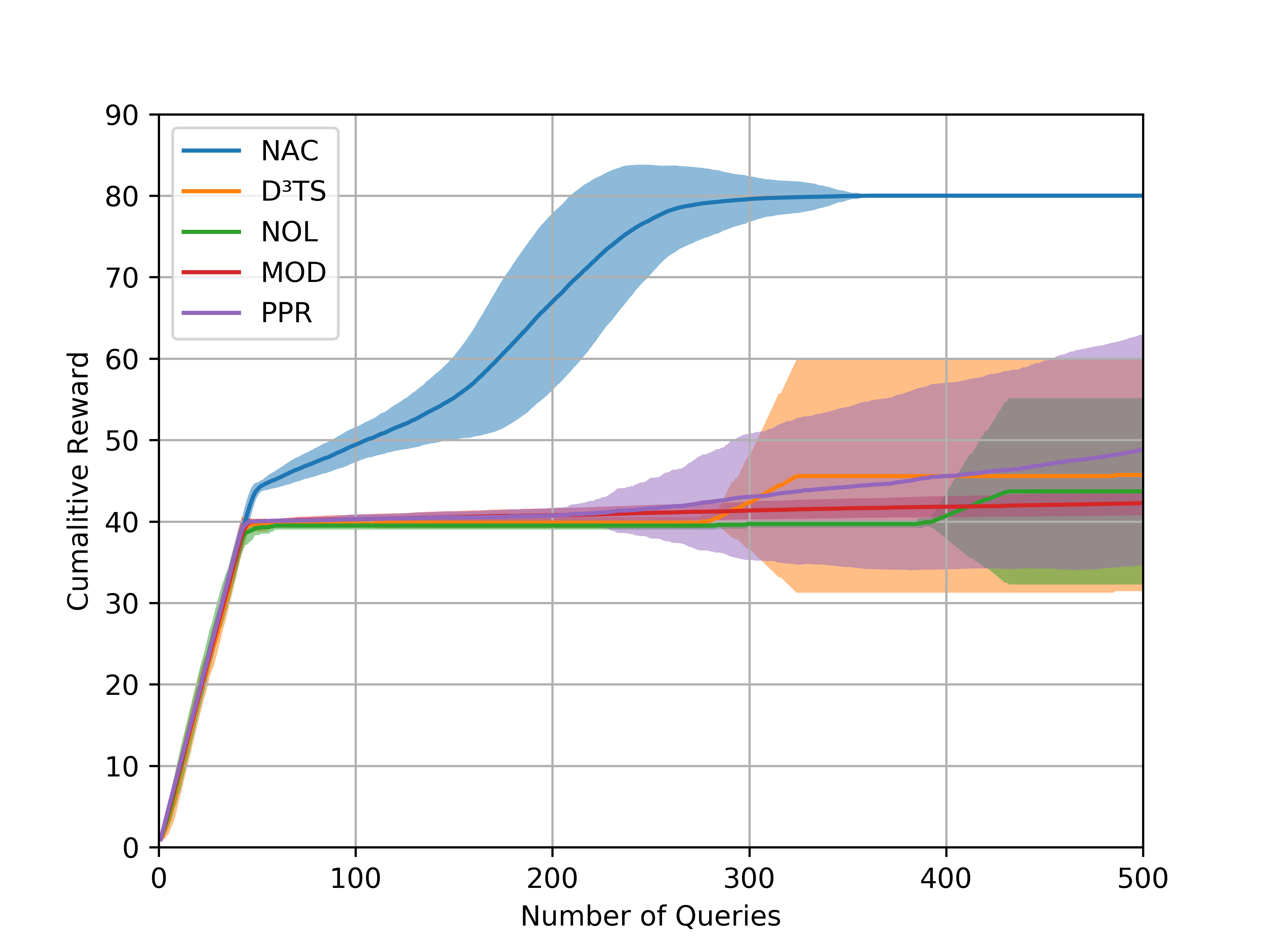}
}

\caption{NAC discovers two anomalous cliques that are not adjacent.}
\label{fig:2-clique}
\end{center}
\end{figure} 

\subsection{Learning Scenarios}
 In the first learning scenario, the goal is to detect a set of distributed anomalous vertices. They are represented by two cliques, each containing 40 vertices, that are embedded 2 to 3 hops away from each other. The training instances are networks generated by the SBM model, while the test cases are network instances generated by the LFR model. In this scenario, the discovery agent has to figure out 1) how to value longer exploration paths over the cost of including nodes not in target set, and 2) how to adjust to topological differences between training and testing instances. In Figure~\ref{fig:2-clique}(a), we consider a test case where detactability of the two cliques with complete network information is relatively easy (average background density where the cliques are embedded is comparatively low). We observe that all the methods are able to find the first clique, yet all the baselines struggle once they enter the region where no clique nodes are present. The baselines eventually find some clique nodes, but, even then, they are unable to fully retrieve the second clique. NAC is able to leverage estimation of long-term reward and access to the offline policy to fully recover both cliques, and furthermore, is able to generalize to the more complex LFR topology.
 
\begin{figure}[ht]
\subfloat[Facebook Politician \label{fig:fb-p}]{
\includegraphics[width=0.32\linewidth]{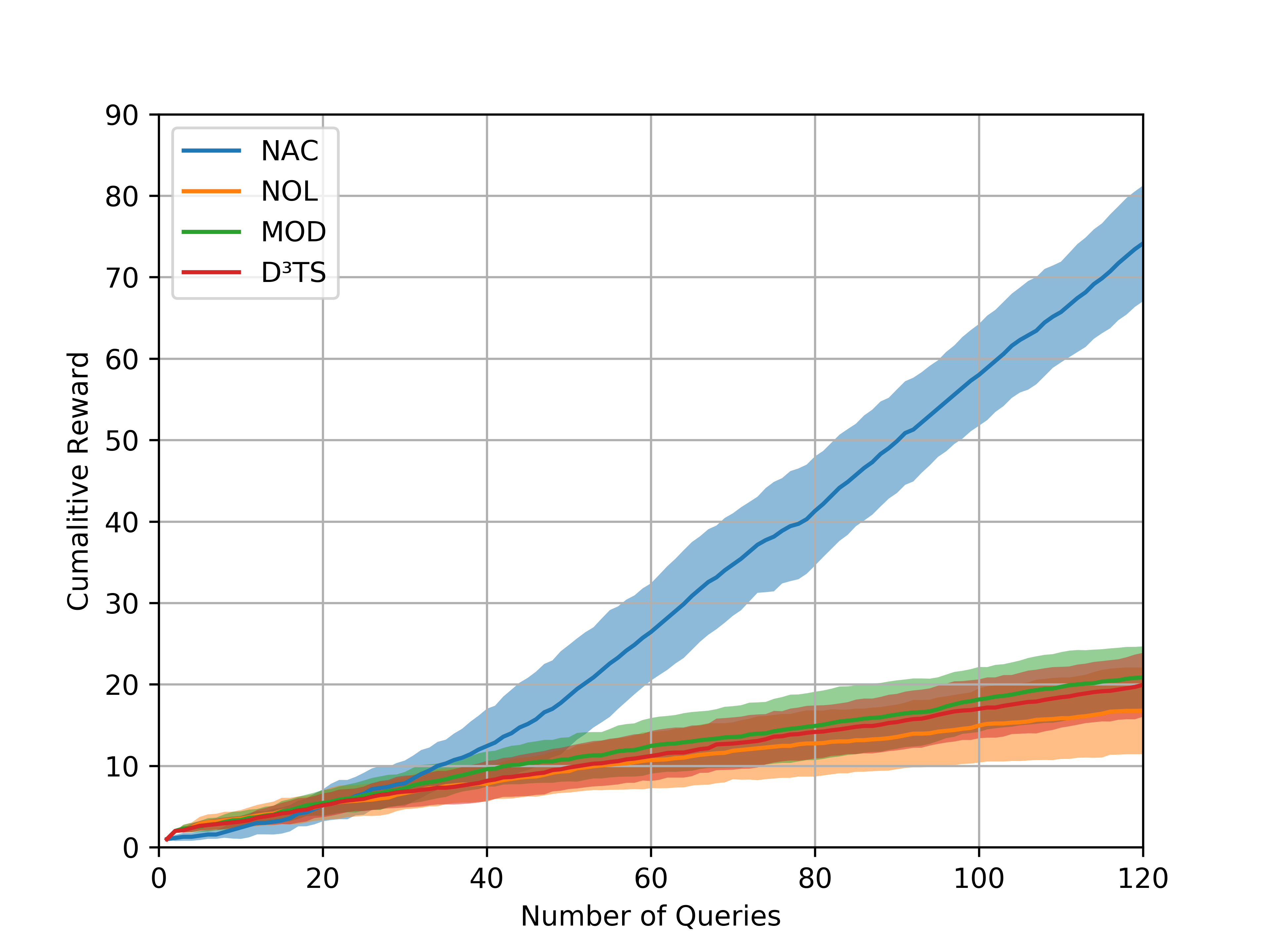}}
\subfloat[Facebook TV Shows \label{fig:fb-tv}]{
\includegraphics[width=0.32\linewidth]{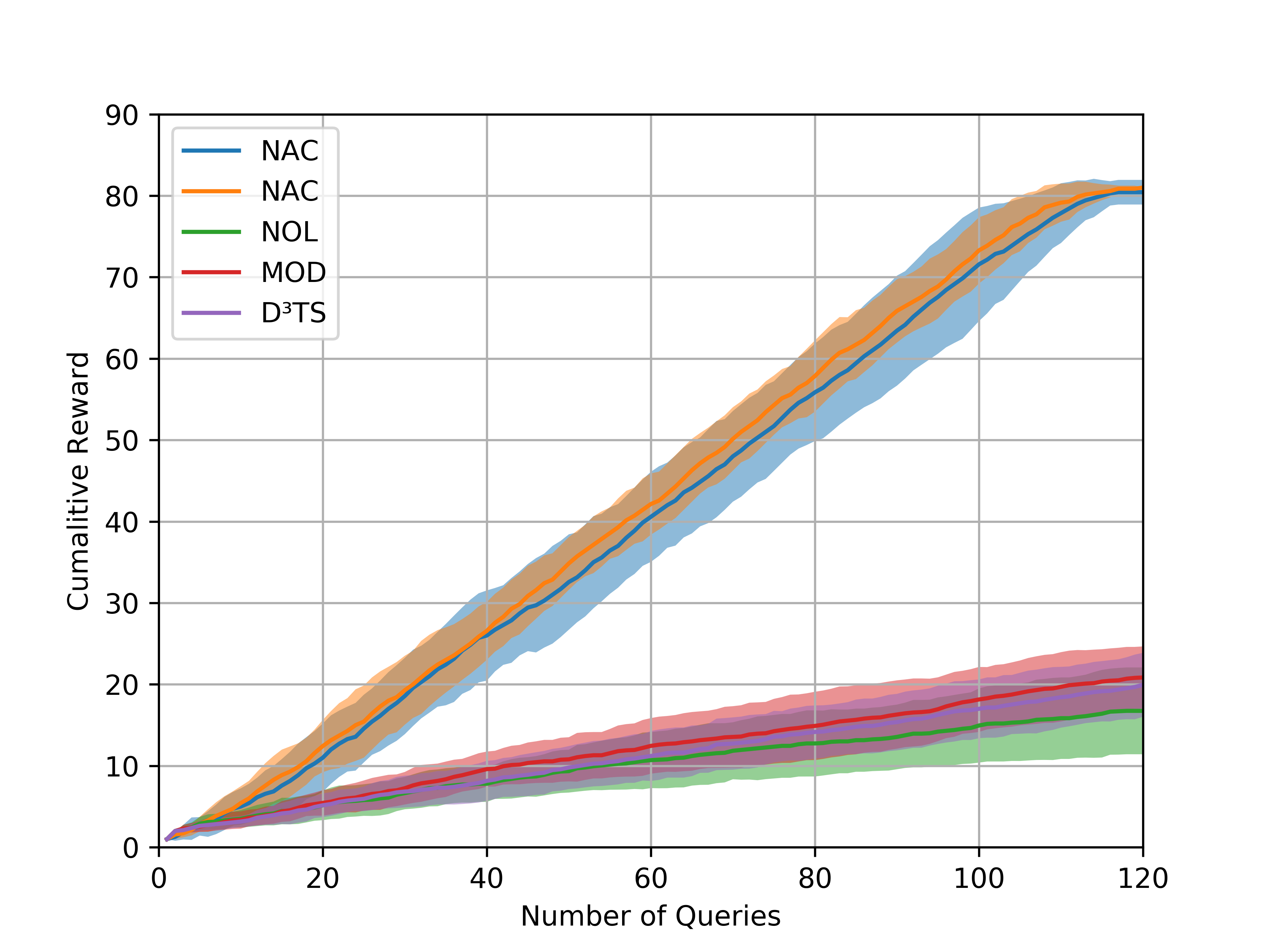}}
\subfloat[Livejournal \label{fig:lj}]{
\includegraphics[width=0.32\linewidth]{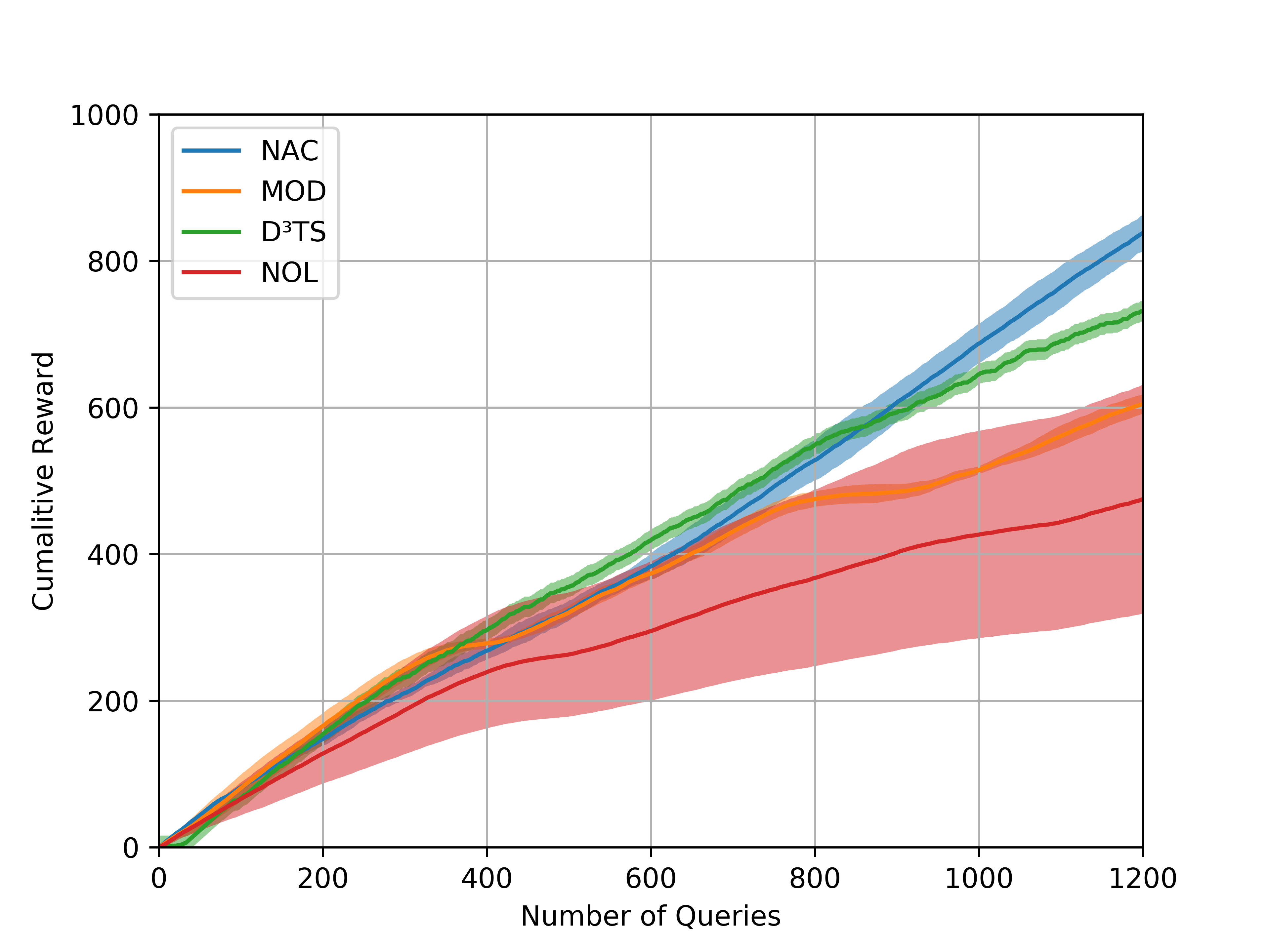}}
\caption{NAC outperforms competitive online methods on real networks.}
\label{fig:real}
\end{figure} 

 In Figure~\ref{fig:2-clique}(a), we consider a much harder case: embedding two disjoint dense subgraphs, each with density 0.2 in a background of density 0.05. These parameters are close to the detectability bound~\cite{Nadakuditi2012} for the complete network case. In this case, neither of the baselines learns how to recover the second clique. NAC goes through a longer exploration phase, but eventually learns how to grow the network to identify the second clique.
 In Figure~\ref{fig:real}(a) and (b), we illustrate how our model trained on synthetic background networks generalizes to realistic background topologies. For this scenario, we trained with instances from both the LFR and SBM models. We observe that NAC generalizes very well to the Facebook network topologies and is able to fully discover the target nodes. 

\section{Role of Network Embedding}
\label{sec:embed}

\begin{figure}[!ht]
\begin{center}
\subfloat[No embedding \label{fig:no-ppr}]{\includegraphics[width=0.320\linewidth]{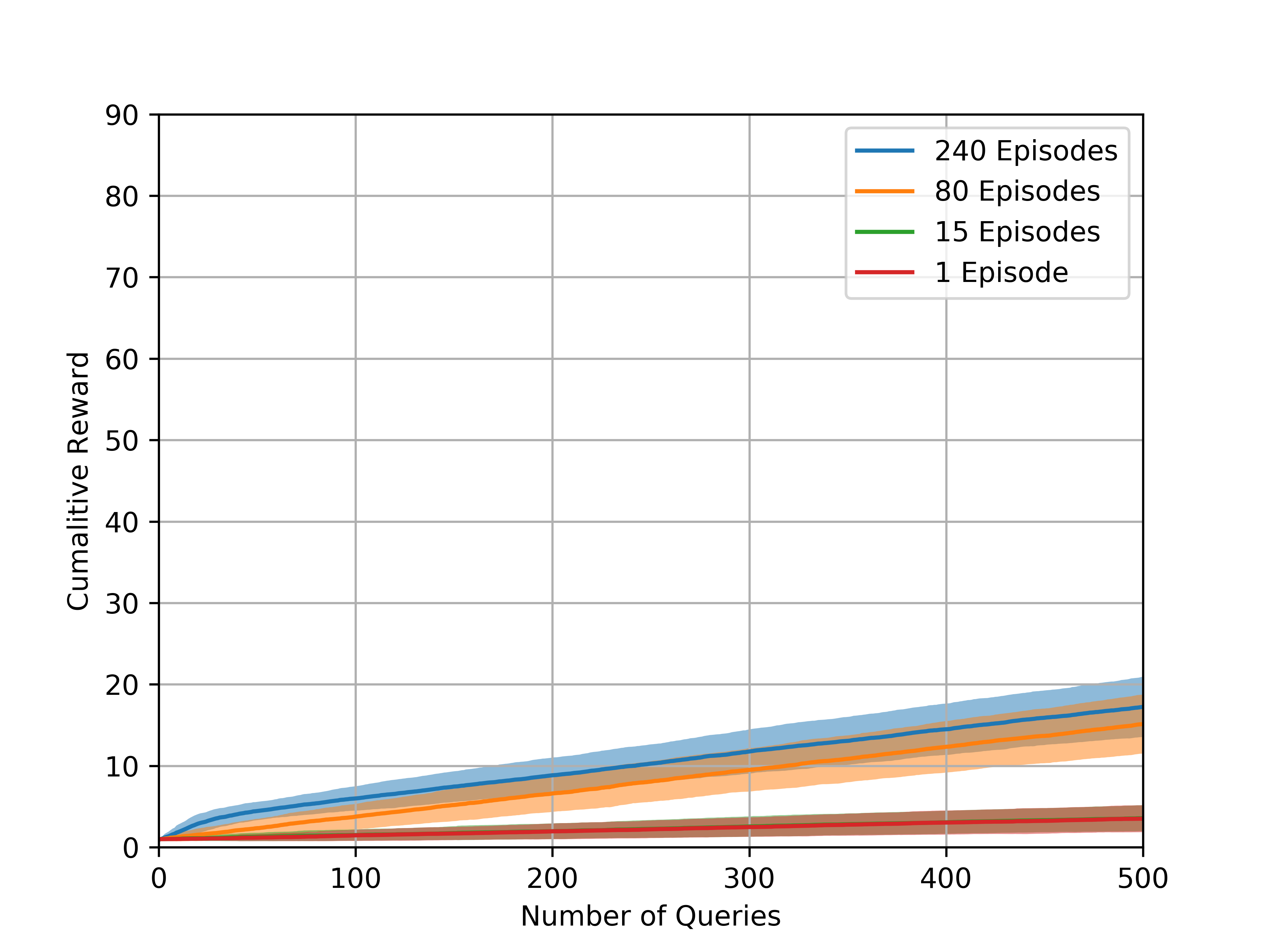}}
\subfloat[With Eigenmap \label{fig:ppr}]{\includegraphics[width=0.320\linewidth]{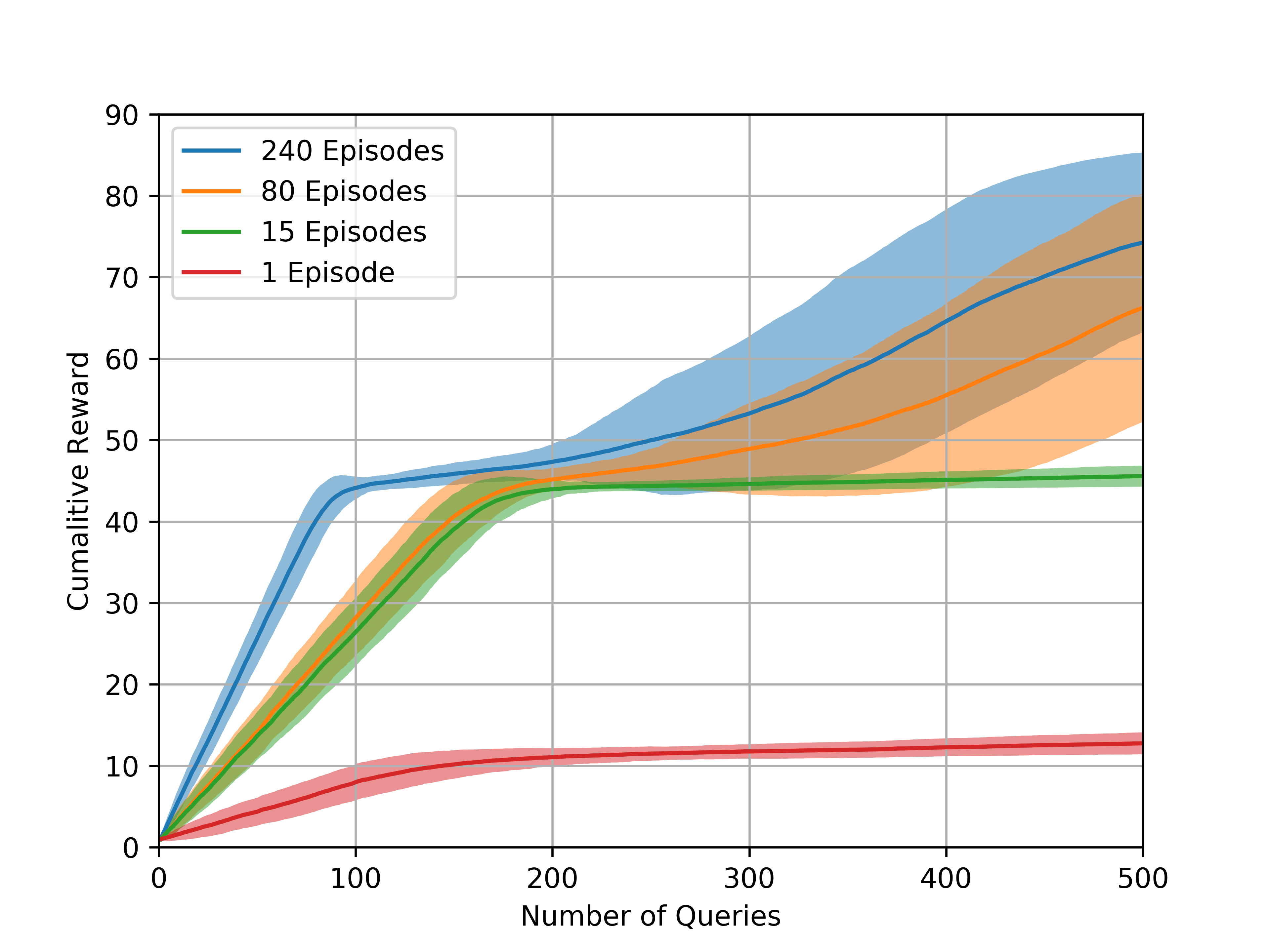}}
\subfloat[With PPR \label{fig:nac-ppr}]{\includegraphics[width=0.320\linewidth]{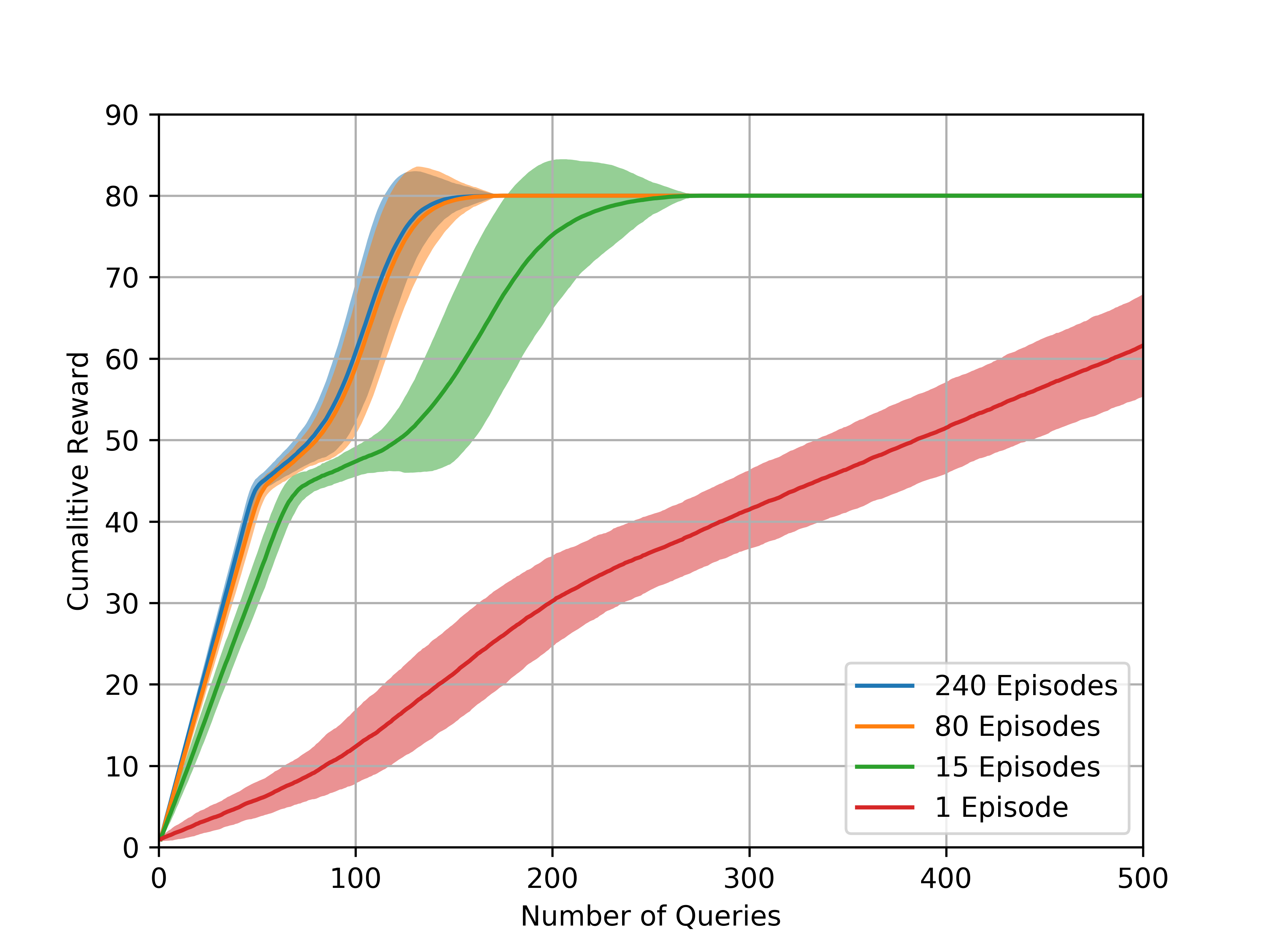}}
\caption{NAC convergence behaviour with and without embedding.}
\label{fig:conv}
\end{center}
\end{figure} 

In this section, we systematically explore the role of the embedding algorithm  in supporting better network discovery for selective harvesting. We consider two broad classes of embedding methods: walk-based methods~\cite{Groover2016,Murai2017,Haveliwala2003}, and matrix factorization methods~\cite{Pearson1901,Torres2019,Belkin2003}.

As introduced earlier, Maximum Degree (MOD)~\cite{Murai2017} is a heuristic embedding which ranks nodes by the number of edges shared with a target node. 
Personalized Page Rank (PPR)~\cite{Haveliwala2003} is a random walk method which ranks nodes by their estimated random walk distance to an observed target nodes. 
We used a damping parameter $\alpha=0.8$.
Node2Vec\cite{Groover2016} is a deep-walk based method which attempts to learn a neighborhood preserving representation for each node in a given graph. We ranked embedded nodes by estimating the Euclidean distance between each node and the observed target nodes. For node2vec, we aggregated estimates over 5 walks of length 40 and embedding dimension of 64. 
Principal Component Analysis~\cite{Pearson1901} computes the eigen-decomposition of the input adjacency matrix. We estimate the node ranking by looking at the average Euclidean distance between a node and observed target nodes. 

Laplace Eigenmap Embedding (Eigenmap)\cite{Belkin2003} is a low dimensional graph representation based on spectral properties of the Laplacian matrix of a graph. In this embedding, we represent vertices using the eigenvector corresponding to the lowest eigenvalue. Rank is estimated by looking at the absolute value of the dot product as described in \cite{Torres2019}. The embedding dimension was set to 64.

Geometric Laplacian Eigenmap Embedding (GLEE)\cite{Torres2019} is a low dimensional graph representation based on geometric properties of the Laplacian matrix of a graph. Unlike eigenmap, GLEE represents vertices using the eigenvector corresponding to the largest eigenvalue. Ranking is estimated in the same way as Eigenmap. The embedding dimension was set to 64.

\subsection{Embedding Evaluation Metrics}
Our evaluation of the embedding algorithm is in context of its support to NAC's policy learning component. An effective RL agent for selective harvesting would benefit from state approximations that reflect canonical states for this task (illustrated in Figure~\ref{fig:canonical-cases}). This may imply differing embedding objectives than if we analyze network embedding algorithms as standalone solutions.
To this effect, we consider the following metrics for evaluating the role of the embedding algorithm.

\paragraph{Consistent embedding:} ideally, we would like the embedding algorithm to place probed and unprobed target nodes near each other. This property implies that NAC will have a higher chance of visiting target nodes earlier than background nodes. To capture the consistency property of the embedding algorithm $e(\cdot)$, we measure, at every discovery step $t$, the accuracy of the embedding algorithm in recovering the top $k$ target nodes:
\begin{equation}
    Accuracy_t(e(G_t)) = \frac{\# \text{top k target nodes identified by embedding}}{\# \text{true target nodes}}.
\end{equation}

\paragraph{Compressability of state-action space:} in an ideal RL setting, the highest reward value $Q(s,a)$ for a given action space, will be highly concentrated over the best action option. We can conceptualize this scenario using a Gaussian distribution with mean represented by the reward value of the best action and minimal variance. We favor embeddings which concentrate favorable actions, e.g. tightly cluster target nodes in the border set.

\begin{figure}[ht]
\begin{center}
\includegraphics[width=0.75\textwidth]{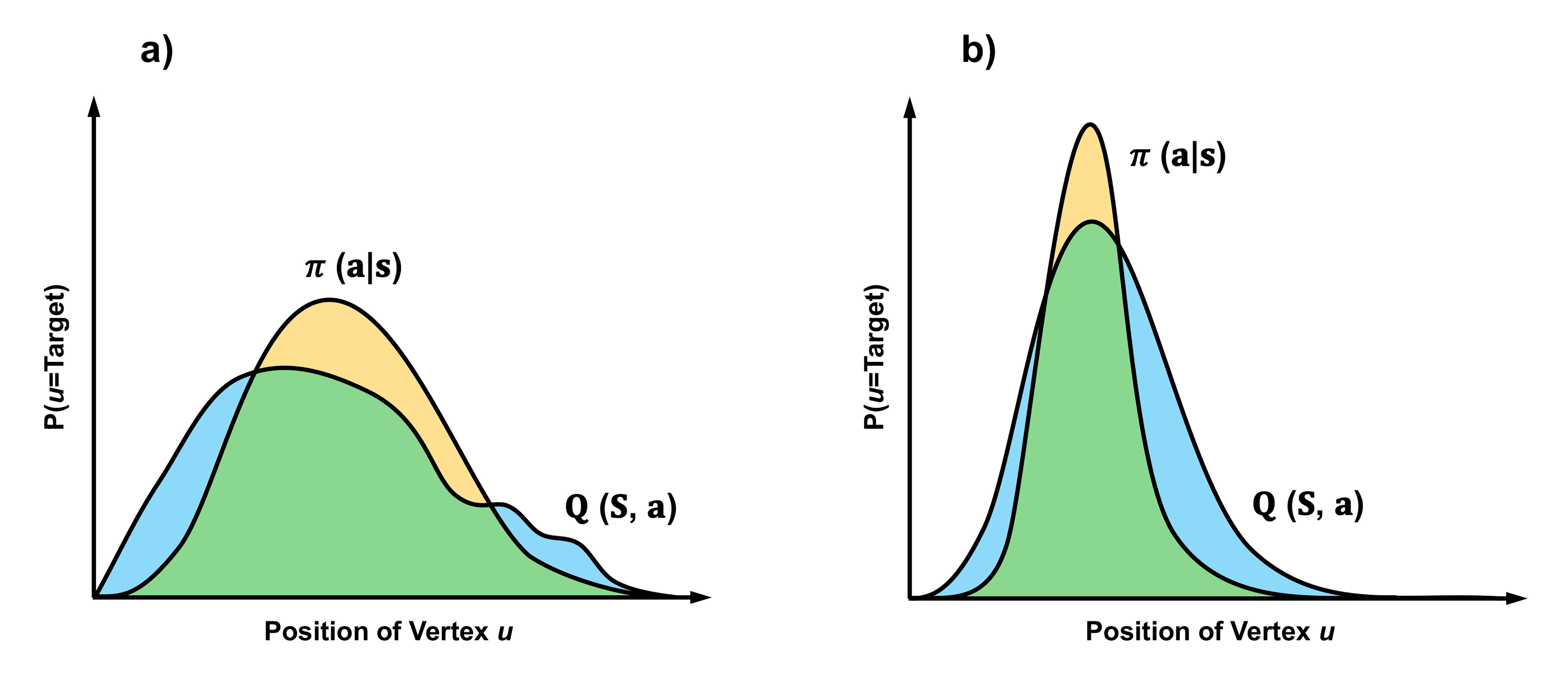}
\caption{Illustration of how quality of node embedding affects the quality of policy and reward functions; a) highly uncertain reward and policy functions b) highly concentrated reward and policy functions. Ideally, our policy is distributed around the best action with minimal variance, so b) is preferred.}
\label{fig:doodle}
\end{center}
\end{figure}

This entropy minimization concept is illustrated by Figure  \ref{fig:doodle}. In~\ref{fig:doodle}a) the entropy for $Q(s=G_t,a=u)$ is higher than \ref{fig:doodle}b) causing the policy $\pi(a=u|s=G_t)$ to have higher variance and lower probability of successfully selecting the ``best" node.

To measure how the embedding algorithm supports this entropy minimization principle, we look at the variance for node rankings in the embedding space for each target cluster and compute the entropy as follows,  
\begin{equation}
    H_{\mathcal{B}_t}(e) = 0.5 [1+\text{log}(2\pi \text{Var}[e(\mathcal{B}_t)])],
\end{equation}
where $\mathcal{B}_t$ is the set of target nodes in the border set.


\paragraph{Robustness to increasing signal complexity:} a good embedding algorithm allows the discovery agent to stay robust as the strength of the signal deteriorates and its complexity increases. We consider two parameters: the strength of background class and the strength of target class and vary them to explore both regions of high and low SNR. To capture robustness, we aggregate the accuracy metric at every discovery step $t$ to compute the Area Under the Curve (AUC),
\begin{equation}
    AUC = \sum_t^{steps}Accuracy_t(e(G_t)),
    \end{equation}
and examine its sensitivity to target and background model parameters.

\paragraph{Learning convergence time:} 
a useful embedding algorithm reduces the number of episodes required to learn effective discovery policies. The embedding algorithm does this by mapping many possible network states to fewer canonical representations that aid policy learning. Here we estimate improvements in NAC's convergence rates without and with access to the embedding step.

\subsection{Data Generation for Embedding Analysis}
\label{sec:data_embed}
 We consider the following learning setting for our embedding analysis: the target class is a set of disjoint dense subgraphs embedded within a background network. A variety of background and anomaly densities are tested. For each learning step the incomplete graph is embedded and a ranking for all observed nodes is computed and scored. An optimal policy is defined as navigating each step of the selective harvesting task in the minimal number of steps. We utilize ground truth to navigate the graph optimally. For illustration, in the setting of two anomalies with 40 nodes each, separated by 2-hops, a perfect traversal is 81 steps long. 

In our experiments, we consider the following parameters: each anomaly has 40 vertices, and the background network consists of 2000 vertices. We use the stochastic block model to generate background instances at various densities. Each background instance contains two communities with intra-community edge probability $p_1=p_2=0.25$ and inter-community edge probabilities $r$ in the range $\{0.01, 0.025, 0.05, 0.075, 0.1\}$.
We use the ER model to generate anomalies with edge probabilities $p_t$ in the range $\{0.25,0.5,0.75,1\}$. 
For each unique set of parameters, we generate 10 graph instances leading to a total of 200 graph instances.

\subsection{Empirical Analysis of Embedding Algorithms}

In this section we summarize our empirical evaluation of a few embedding algorithms.

\begin{figure}[!htb]
\begin{center}
\includegraphics[width=0.85\textwidth]{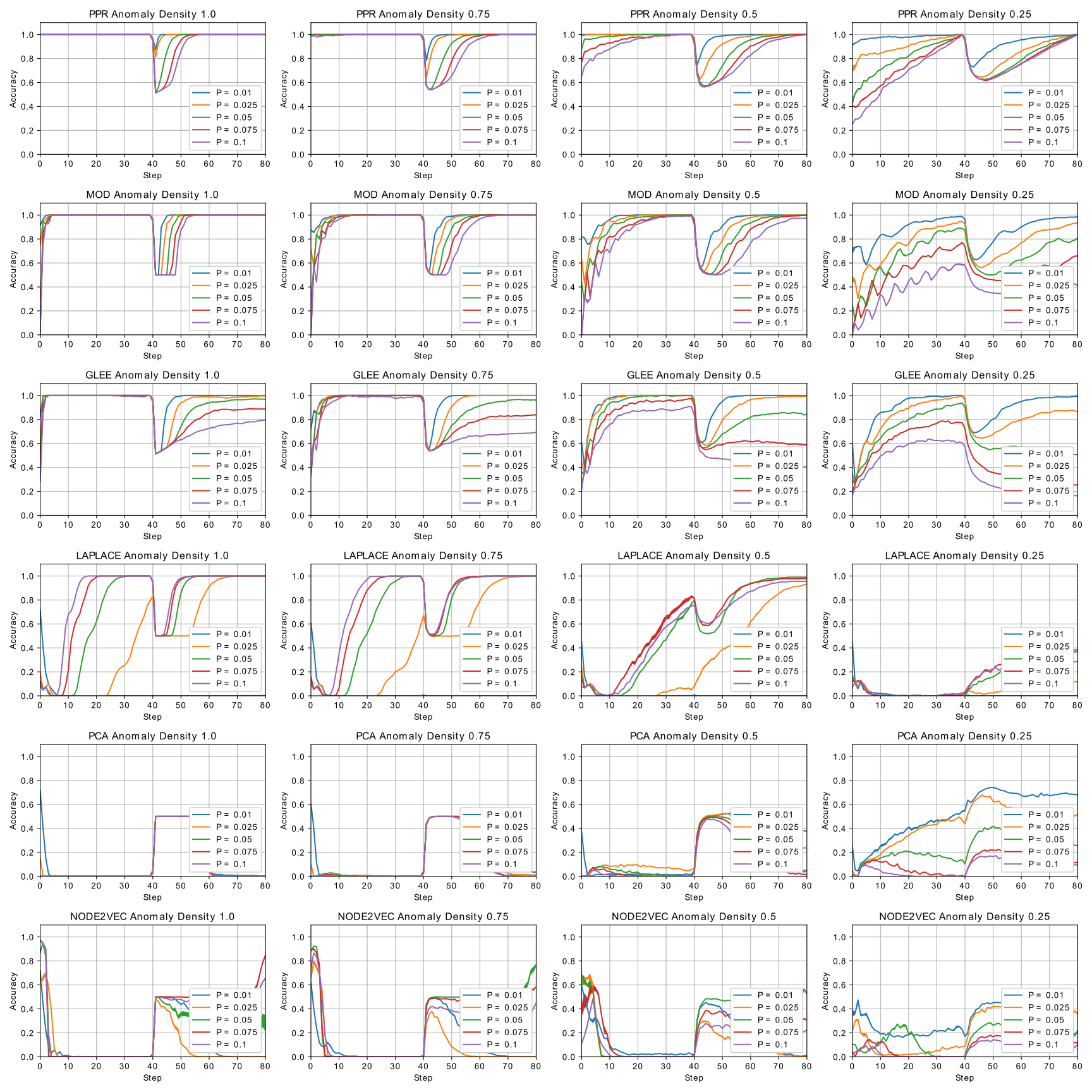}
\caption{Accuracy index vs time step for all tested embeddings and configurations. The results are arranged from top to bottom by the embedding type: PPR, MOD, GLEE, Laplacian, PCA, and Node2Vec. The anomalies implanted in the dataset get sparser from left to right, causing the task to increase in complexity and in general for the performance of each embedding to diminish. We observe that PPR, MOD, GLEE, and Laplacian embedding all perform well on the task and PPR maintaining the best performance.}
\label{fig:jaccard}
\end{center}
\end{figure} 

\paragraph{Consistent embedding} property
is analyzed in Figure~\ref{fig:jaccard}.
Within the two target regions, we observe that walk-based methods, PPR and MOD, do a fairly good job in prioritizing target nodes for subsequent selection. PPR is much more robust as the strength of the anomaly weakens relative to the background network. Node2vec method by contrast struggles in the same regions, though it seems to do slightly better once the agent discovers the second clique. It is possible that increasing the dimensionality of the feature vectors would lead to improved performance, however this method is computationally intensive as we consider emebeddings over many learning iterations and many graph instances.


Overall, across all the embeddings, we observe a strong drop in performance when transitioning from what should be the exploitation and exploration regimes. We observe similar embedding sensitivity to decreasing levels of SNR. These observations highlight the role the offline policy learning is playing in recognizing and adapting to changing discovery regimes and sparse task-related signals. 

\begin{figure}[!htb]
\begin{center}
\includegraphics[width=0.85\textwidth]{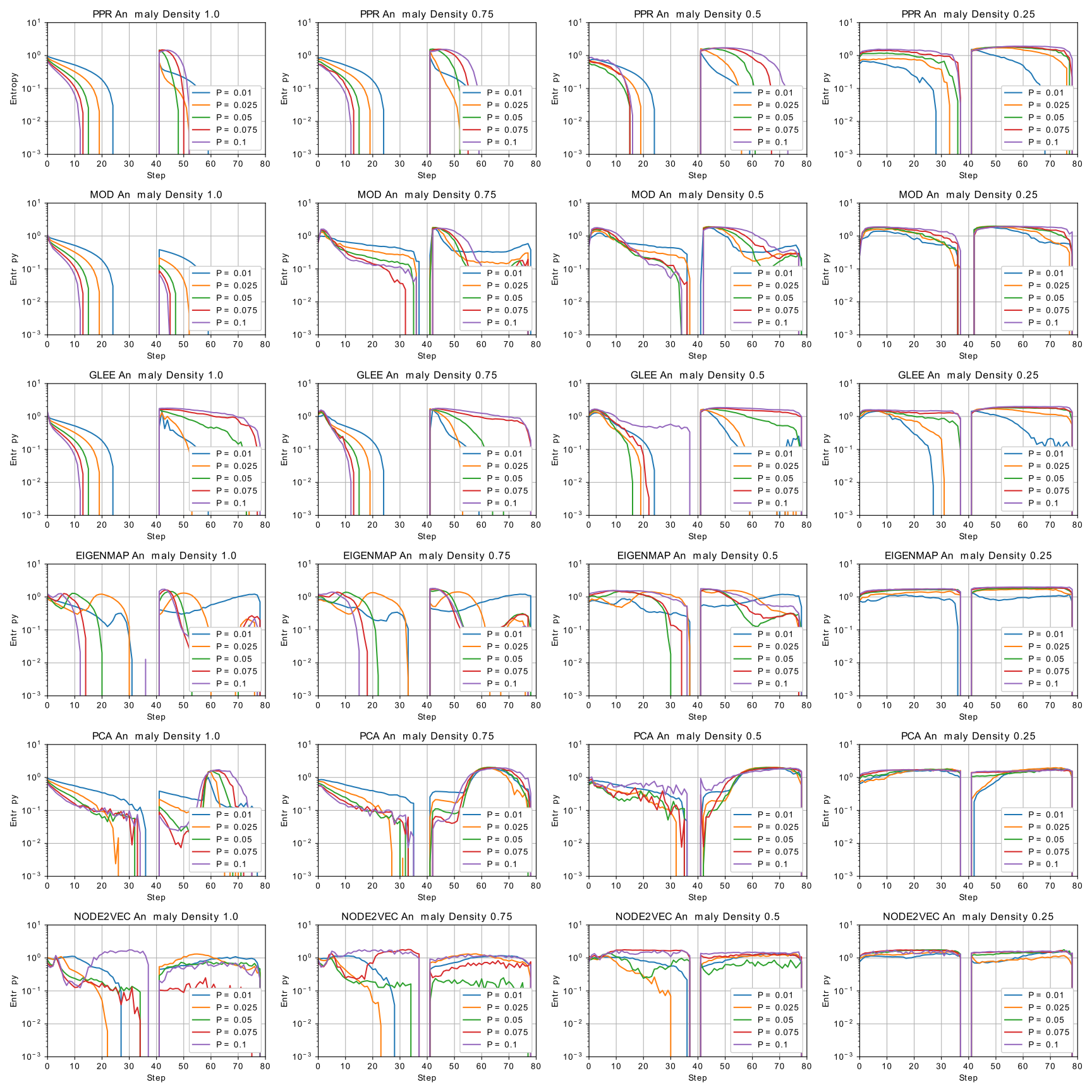}
\caption{Entropy vs time step for all tested embeddings and configurations. Here we see that MOD, PPR, LAPLACE, and GLEE are able to compress the action space, which is indicated by the sharp drops in entropy before the entire anomaly has been discovered ($<$40 steps). This is especially pronounced in the easier cases, e.g. denser anomalies, illustrated in the leftmost plots for each embedding.}
\label{fig:entropy}
\end{center}
\end{figure} 

\paragraph{Compressability of state-action space}
is illustrated in Figure~\ref{fig:entropy}. Similar to the accuracy metric, PPR and MOD appear to do the best job of quickly collapsing to a set of vertex positions as enough target nodes are collected. Again, node2vec appears to be a poor choice, but does exhibit some compressability in the higher SNR cases. The graph factorization approaches appear to follow the expected trend of degrading in performance with a reduction of SNR. 
\begin{figure}[!htb]
\begin{center}
\includegraphics[width=0.85\textwidth]{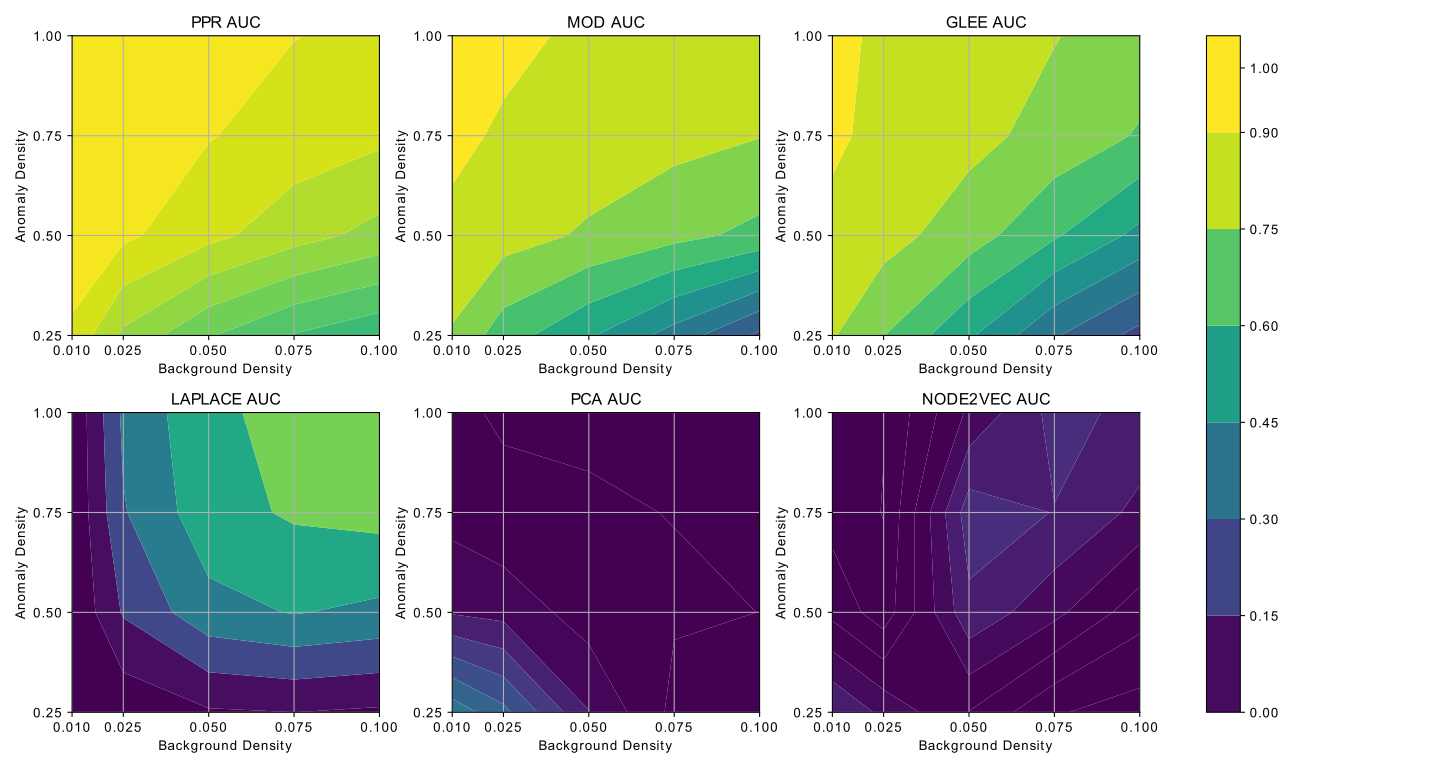}
\caption{AUC vs background and anomaly density for all tested embeddings. In general, we see that PPR performs best in all scenarios presented in Section \ref{sec:data_embed}.}
\label{fig:auc}
\end{center}
\end{figure} 

\paragraph{Robustness to increasing signal complexity} is demonstrated by Figure \ref{fig:auc}, which represents the integrated accuracy over the entire select harvesting task. The same trends discussed in the accuracy section are illustrated here. This figure delineates, at an aggregate level, the network topology characteristics where simple embedding heuristics are sufficient to support effective selective harvesting (lighter color regions) and those topology characteristics where offline planning is required.

\paragraph{Learning convergence time} is analyzed in Figure~\ref{fig:conv}, which shows the performance of each embedding paired with policy learning after N episodes of training. We demonstrate in Figure~\ref{fig:conv}a) that without embedding, the convergence time is likely to be very long and requires a high capacity network. We show in Figure~\ref{fig:conv}b) and c) representative embeddings from the walk-based and factorization classes and observe that they converge in a consistent way to their entropy and accuracy scores. We illustrate by analyzing the test case described in Figure~\ref{fig:2-clique}(a), but the behavior is consistent for all the different test cases considered. 
Overall we observe that embedding quality directly impacts convergence time and ultimately the ability of the discovery algorithm to achieve the downstream task objective with budget and resource constraints. Across the various evaluation metrics and learning regimes, we consistently observe PPR outperforming other embedding algorithms in best augmenting discovery policy learning for selective harvesting.
The success of PPR across the various evaluation metrics could be explained by the shared characteristics between the selective harvesting task and the PPR algorithm. Both algorithms rely on the concept of exploring local, relatively dense neighborhoods from a seed node. The same rationale can explain the relative success of the MOD heuristic, though MOD does not have the randomness feature that allows PPR to handle sparser distributions of target nodes. The rest of the embedding approaches lack the seed-centric embedding property and therefore never match the overall performance of PPR. Our hypothesis, however, is that consideration of alternative downstream tasks, might imply a different ranking of suitable embedding methods.

\section{Conclusions and Future Work}
We introduced NAC, a deep RL framework for task-driven discovery of incomplete networks. NAC learns offline models of reward and network discovery policies based on a synthetically generated training set. NAC is able to learn effective strategies for the task of selective harvesting, especially for learning scenarios where the target class is relatively small and difficult to discriminate. We show that NAC strategies transfer well to unseen and more complex network topologies including real networks. We analyze various network embedding algorithms as mechanisms for supporting fast navigation through the large network state space. Across several metrics of evaluation, we identify personalized Pagerank as a robust embedding strategy that best supports selective harvesting planning. We leave analysis of alternative downstream tasks and the analysis of respective suitable embeddings for future work.

Our approach opens up many interesting venues for future research. The effectiveness and convergence of our algorithm relies on being trained on a sufficiently representative training set. It is valuable to further explore and quantify the limits of transferability of synthetically generated training sets. Interestingly, our current framework is flexible enough to incorporate additional discovery strategies generated from other methods, as part of the offline training process. This feature can lead to more efficient discovery strategies, but we leave the careful analysis for future work.
Finally, the framework is general enough to support discovery for other network learning tasks. It is valuable to explore how a different learning objective changes the training, convergence, and generalizibility requirements.

\end{document}